\documentclass{article}

     \PassOptionsToPackage{numbers, compress}{natbib}

\usepackage[preprint]{neurips_2020}

\usepackage[utf8]{inputenc} 
\usepackage[T1]{fontenc}    
\usepackage{hyperref}       
\usepackage{url}            
\usepackage{booktabs}       
\usepackage{amsmath, amsfonts}       
\usepackage{nicefrac}       
\usepackage{microtype}      
\usepackage{graphicx}
\usepackage{multirow}       
\usepackage{subcaption}     
\usepackage{bbm}
\usepackage{xcolor}
\usepackage{algpseudocode,algorithm,algorithmicx}

\title{Online Kernel based Generative Adversarial Networks}

\author{
  Yeojoon Youn \\
  Georgia Institute of Technology\\
  Atlanta, GA 30332 \\
  \texttt{yjyoun92@gatech.edu} \\
  \And
  Neil Thistlethwaite \\
  Georgia Institute of Technology \\
  Atlanta, GA 30332\\
  \texttt{nthistle@gatech.edu} \\
  \AND
  Sang Keun Choe \\
  Carnegie Mellon University \\
  Pittsburgh, PA 15213 \\
  \texttt{sangkeuc@andrew.cmu.edu} \\
  \And
  Jacob Abernethy \\
  Georgia Institute of Technology \\
  Atlanta, GA 30332 \\
  \texttt{prof@gatech.edu} \\
}

\definecolor{red}{RGB}{255,00,00}

\usepackage{color}              
\usepackage{color-edits}
\addauthor{yeojoon}{cyan}
\addauthor{sangkeun}{red}
\addauthor{neil}{blue}
\addauthor{jake}{magenta}

\begin{document}

\maketitle

\begin{abstract}
One of the major breakthroughs in deep learning over the past five years has been the Generative Adversarial Network (GAN), a neural network-based generative model which aims to mimic some underlying distribution given a dataset of samples. In contrast to many supervised problems, where one tries to minimize a simple objective function of the parameters, GAN training is formulated as a min-max problem over a pair of network parameters. While empirically GANs have shown impressive success in several domains, researchers have been puzzled by unusual training behavior, including cycling so-called \emph{mode collapse}. In this paper, we begin by providing a quantitative method to explore some of the challenges in GAN training, and we show empirically how this relates fundamentally to the parametric nature of the discriminator network. We propose a novel approach that resolves many of these issues by relying on a kernel-based non-parametric discriminator that is highly amenable to online training---we call this the Online Kernel-based Generative Adversarial Networks (OKGAN). We show empirically that OKGANs mitigate a number of training issues, including mode collapse and cycling, and are much more amenable to theoretical guarantees. OKGANs empirically perform dramatically better, with respect to reverse KL-divergence, than other GAN formulations on synthetic data; on classical vision datasets such as MNIST, SVHN, and CelebA, show comparable performance.
\end{abstract}

\section{Introduction}

Generative Adversarial Networks (GANs) \cite{goodfellow2014generative} frame the task of estimating a generative model as solving a particular two-player zero-sum game. One player, the generator $G$, seeks to produce samples which are indistinguishable from those generated from some true distribution $p_{\text{real}}$, and the other player, the discriminator $D$, aims to actually distinguish between such samples. In the classical setting each of these players is given by neural networks parameterized by $\theta_g$ and $\theta_d$, respectively. For a random input seed vector $z$, the generator outputs a synthetic example $x = G(z;\theta_g)$, and the discriminator returns a probability (or score) according to whether the sample is genuine or not. In its original formulation, a GAN is trained by solving a min-max problem over the parameters $\theta_g$ and $\theta_d$:
\begin{gather} \label{eq:ganformulation}
    \min_{\theta_g} \max_{\theta_d} \Big\{ V(\theta_d, \theta_g) = \mathbb{E}_{x \sim p_{\textrm{data}}}[\log D_{\theta_d}(x)] + \mathbb{E}_{z \sim p_z}[\log (1-D_{\theta_d}(G_{\theta_g}(z)))] \Big\}
\end{gather}
From a theoretical perspective, this framework for learning generative models has two very appealing qualities. First, in a setting where $V$ is convex in $\theta_g$ and concave in $\theta_d$, a true equilibrium point of this game could be readily obtained using various descent-style methods or regret-minimization \cite{kodali2017convergence}. Second, it was observed in \cite{goodfellow2014generative} that, given an infinite amount of training data, and optimizing over the space of all possible discriminators and all possible generative models, the equilibrium solution of \eqref{eq:ganformulation} would indeed return a generative model that captures the true distribution $p_{\text{real}}$. \cite{zhao2016energy, mao2017least} also claim that their GANs can learn the true distribution based on this strong assumption.

The challenge, in practice, is that none of these assumptions hold, 
at least for the way that the most popular GANs are implemented. The standard protocol for GAN training is to find an equilibrium of \eqref{eq:ganformulation} by alternately updating $\theta_d$ and $\theta_g$ via stochastic gradient descent/ascent using samples drawn from both the true distribution (dataset) and generated samples. It has been observed that simultaneous descent/ascent procedures can fail to find equilibria even in convex settings \cite{mescheder2017numerics, mescheder2018training, daskalakis2017training}, as the equilibria are ``unstable'' when traversal through the parameter landscape is viewed as a dynamical system; one might even expect to see cycling around the min-max point without progress.

Arora et al. \cite{arora2017generalization} raise another issue
involving the capacity of the discriminator network. In the thought experiment from \cite{goodfellow2014generative} we summarized above, the  discriminator must have ``enough capacity'' to produce a suitably complex function. In practice, however, GANs are trained with discriminators from a parametric family of neural networks with a fixed number of parameters $k$. \cite{arora2017generalization} observes that if the generator is allowed to produce distributions from mixtures of $O\left(\frac{k}{\epsilon^2}\right)$ base measures then the generator can fool the discriminator (up to $\epsilon$) with the appropriate choice of mixture that differs substantially from $p_{\text{real}}$. Authors of this work have suggested that this issue of finite capacity explains GAN \emph{mode collapse}, a phenomenon observed by practitioners whereby a GAN generator, when trained to estimate multi-modal distributions, ends up dropping several modes.

We would argue that the problem of mode collapse and ``cycling'' are strongly related, and arise fundamentally from the parametric nature of the discriminator class of most GANs. Here is an intuitive explanation for this relationship: One can view the interaction between the two opponents as $D$ playing a game of \textit{whack-a-mole} against $G$. As $G$ is pushed around by the discriminating $D$, modes can appear inadvertently in the training process of $G$ that $D$ had not previously considered. With limited capacity, $D$ will have to drop its discrimination focus on previously ``hot'' regions of the input space and consider these new bugs in the generator. But now that $D$ has lots of focus on such regions, $G$ can return to producing new modes here, and $D$ will again have to reverse course. 

\begin{figure}[!htbp]
    \centering
    \begin{subfigure}[b]{0.35\textwidth}
    \includegraphics[width=\textwidth]{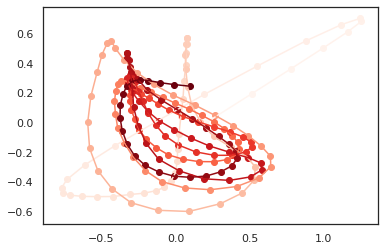}
    \end{subfigure}
    \quad
    \begin{subfigure}[b]{0.35\textwidth}
    \includegraphics[width=\textwidth]{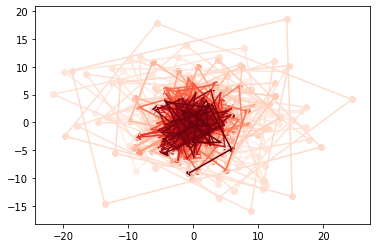}
    \end{subfigure}
    \caption{\footnotesize
    Qualitative comparison of Vanilla GAN (Left) and OKGAN (Right, proposed) on
    cycling behavior on 2D-grid dataset (see Section \ref{experimental_setup}). 
    Vanilla GAN shows cycling behavior; 
    the parameters of discriminator cycle around the equilibrium, slowing down the convergence to the optimum and causing the generator to be unable to effectively learn the real distribution.
    In contrast, OKGAN does not appear to suffer from cycling behavior.
    }
    \label{cycling_graph}
\end{figure}

We have found a useful way to visualize the dance performed between $D$ and $G$. While it is hard to observe cycling behavior in the high-dimensional parameter space of $\theta_d$, we can take a natural projection of the discriminator as follows. Let $\theta_d^{(1)}, \theta_d^{(1)}, \ldots, \theta_d^{(T)}$ be the sequence of parameters produced in training the GAN via the standard sequential update steps on the objective \eqref{eq:ganformulation}, and let $x_1, \ldots, x_m$ be a random sample of examples from the training set. Consider the matrix $M = [D(x_j; \theta_d^{(i)})]_{i \in [T], j \in [m]}$, where each row represents the function value of the discriminator on a set of key points at a fixed training time $i$. We can perform a Principle Component Analysis of this data, with two components, and obtain a projected version of the data $\tilde M \in \mathbb{R}^{T \times 2}$, so each row is a 2-dimensional projection of the discriminator at a given time. We display this data in Figure~\ref{cycling_graph} (Left graph), where light pink nodes are earlier training time points and dark red are later rounds of training. The observation here is quite stark: the discriminator does indeed cycle in a relatively consistent way, and suggests that training may not always be highly productive.

In the present paper, we propose a GAN formulation that relies on a \emph{non-parametric} family of discriminators, kernel-based classifiers, which can be efficiently trained online; we call this OKGAN (the Online Kernel-based Generative Adversarial Network). OKGANs exhibit a several benefits over previous GAN formulations, both theoretical and empirical. On the theory side, the non-parametric nature of the family of functions helps avoid the two negative results we highlighted above: the limited-capacity challenge raised by \cite{arora2017generalization} and the numerical training issues described by \cite{mescheder2017numerics}. Given that the kernel classifiers can grow in complexity with additional data, the discriminator is able to adapt to the increasing complexity of the generator and the underlying distribution to be learned\footnote{Our kernel classifiers do require a limited budget size for the number of examples to store, but this is mostly for computational reasons, and the budget size can be scaled up as needed}.  Furthermore, the discriminator formulation is now a convex problem, and we have a wealth of results on both the computational as well as statistical properties of kernel-based predictors. Kernel classifiers also allow additional flexibility on data representation through the choice of kernel, or combination of kernels. For example, we have found that mixtures of gaussian kernels with different radii perform best in complicated image datasets such as CelebA.

What is quite clear, empirically, is that OKGANs \textit{do not suffer from mode collapse}, at least not on any of the low-dimensional synthetic examples that we have tested. On simple problems where the target exhibits a density function, we show OKGAN dramatically outperforms prior methods when it comes to capturing the target distribution via reverse KL-divergence. We show that OKGANs achieve the highest diversity on synthetic datasets by using quantitative metrics. Additionally, when we include the use of an ``encoder,'' we qualitatively demonstrate OKGANs work well on classical image datasets, including MNIST, SVHN, and CelebA. We observe the discriminator of OKGANs adapts in a more aggressive fashion, and the discriminator does not appear to exhibit cycling behavior, a common phenomenon for other GANs using neural net based discriminators. 


\subsection{Related works}
\label{related_works}

Many theoretical works have aimed at understanding GANs. \cite{arora2017generalization, zhang2017discrimination} study the generalization properties of GANs under neural distance, \cite{liu2017approximation} studies the convergence properties of GANs via ``adversarial'' divergence. \cite{bai2018approximability} asserts that the diversity in GANs can be improved by letting the discriminator class get strong distinguishing power against a certain generator class. 
\cite{nagarajan2017gradient, mescheder2017numerics, mescheder2018training, li2017limitations, liang2018interaction, nie2019towards} consider a range of questions around GAN dynamics.

Mode collapse and cycling behavior are two of the main issues raised about GAN training. \cite{goodfellow2016nips} observes the cycling behavior of the generator's output in an experiment with a 2D synthetic dataset when mode collapse occurs. \cite{berard2019closer} proposes a new visualization technique called path-angle to study the game vector field of GANs and shows cycling behavior empirically with this technique. \cite{daskalakis2017training} tries to improve GAN training via so-called ``optimistic'' mirror descent.
 \cite{metz2016unrolled} uses an approximately optimal discriminator for the generator update by formulating the generator objective with an unrolled optimization of the discriminator. \cite{srivastava2017veegan} adds a reconstructor network that maps the data distribution to Gaussian noise, providing more useful feature vectors in training. \cite{arora2017theoretical} explores the limitations of encoder-decoder architectures to prevent mode collapse. \cite{lin2018pacgan} proposes a mathematical definition of mode collapse and gives an information-theoretic analysis. Bourgain Theorem mentioned in \cite{xiao2018bourgan} uses metric embeddings to construct a latent Gaussian mixture, a direct approach to solve mode collapse.

Our work is not the first to implement kernel learning ideas into GANs. Perhaps the first such was introduced by \cite{gretton2007kernel}, a statistical hypothesis testing framework called Maximum Mean Discrepancy (MMD), which aims to distinguish between real and fake distributions, rather than explicitly constructing a discriminator. \cite{li2015generative, dziugaite2015training} propose generative moment matching network (GMMN) with a fixed Gaussian kernel for the MMD statistical testing. \cite{li2017mmd} introduces MMD GANs which improve GMMN by adding an injective function to the kernel and making the kernel trainable. \cite{binkowski2018demystifying} demonstrates the superiority of MMD GANs in terms of gradient bias, and \cite{wang2018improving} improves MMD GANs with a repulsive loss function and a bounded Gaussian kernel.

\section{Online kernel GANs}
\label{online_kernel_gans}


\subsection{Online kernel classifier}
\label{online_kernel_classifier}

In the classical formulation of a GAN, the discriminator can generally be regarded as a classifier that aims to distinguish between data in the training set, sampled from $p_{\text{real}}$, and so-called ``fake'' data produced by the generator. The original GAN formulation of \cite{goodfellow2014generative}, and nearly every other generative models inspired by this work \cite{salimans2016improved, nowozin2016f, chen2016infogan, gulrajani2017improved, berthelot2017began, karras2017progressive, miyato2018spectral}, the discriminator is a finitely-parameterized neural network with parameters $\theta_d \in \mathbb{R}^k$. It has generally been believed that the discriminator model family should be suitably complex in order to guide the generator to accurately mimic a complex distribution, and thus a deep neural network was the obvious choice for $D$. What we argue in this paper is that a more classical choice of discriminator model, a function class based on a \textit{Reproducing Kernel Hilbert Space} (RKHS) \cite{bottou2007support}, possesses suitable capacity and has a number of benefits over deep networks. For example, the learning task is indeed a convex problem, which provides guaranteed convergence with well-understood rates. Second, using margin theory and the RKHS-norm to measure function size, we have an efficient way to measure the generalization ability of classifiers selected from an RKHS, and thus to regularize appropriately. Third, they are well suited to fast online training, with regret-based guarantees.

\paragraph{An overview of kernel learning methods.} We now review the basics of kernel-based learning algorithms, and online learning with kernels; see \cite{gretton2007kernel,kivinen2004online,cortes1995support,Dekel2008-qb,Scholkopf2001-xi} for further exposition. Let $\mathcal X$ be some abstract space of data, which typically, although not necessarily, is some finite-dimensional real vector space. A \textit{kernel} $k : \mathcal X \times \mathcal X$ is called \emph{positive semi-definite} if it is symmetric function on pairs of examples from $\mathcal X$, and for every positive integer $d$ and every set of examples $x_1, \ldots, x_d \in \mathcal X$ the matrix $[k(x_i,x_j)]_{i\in[d], j\in[d]}$ is positive semi-definite. Typically we view a PSD kernel as a dot product in some high-dimensional space, and indeed a classic result is that for any PSD $k$ there is an associated feature map $\Phi$ which maps points in $\mathcal X$ to some (possibly infinite dimensional) Hilbert space for which $k(x,x') = \langle \Phi(x), \Phi(x') \rangle$ \cite{Scholkopf2001-xi}. Given a kernel $k$, we can consider functions of the form $f(x) := \sum_{i=1}^m \alpha_i k(x_i, x)$, where the $\alpha_i$'s are arbitrary real coefficients and the $x_i$'s are arbitrary points in $\mathcal X$. The set of functions of this form can be viewed as a \textit{pre Hilbert space}, using the norm $\| f\|_{\mathcal{H}} := \sqrt{\sum_i \sum_j \alpha_i \alpha_j k(x_i, x_j)}$, and when we complete this set of functions we obtain the Reproducing Kernel Hilbert Space $\mathcal{H}$. Again, this is a very brief survey, but more can be found in the excellent book of Sch\"olkopf and Smola \cite{Scholkopf2001-xi}.

Let us give an overview of learning in an RKHS associated to some kernel $k$. First,  imagine we have a sequence of $m$ examples $S_m := \left\{(x_1, y_1), \cdots, (x_m, y_m)\right\}$ sampled from some distribution $p$ on $\mathcal{X} \times \mathcal{Y}$, where $\mathcal{Y}=\{-1, 1\}$. Our goal is to estimate a classifier $f:\mathcal{X} \rightarrow \mathbb{R}$ in $\mathcal{H}$. Assume we have some convex loss function $l:\mathbb{R} \times \mathcal{Y} \rightarrow \mathbb{R}$, where $l(f(x), y)$ is the cost of predicting $f(x)$ when the true label is $y$; typically we will use the \textit{hinge loss} or the \textit{logistic loss}. In a batch setting, we may estimate $f$ given $S_m$ by minimizing the regularized risk functional $R_{\textrm{reg}, \lambda}$ defined as follows: 
\begin{gather} \label{eq:kernel_objective}
    \textstyle R_{\textrm{reg}, \lambda}[f, S_m] := \frac 1 m \sum_{i=1}^m l(f(x_i), y_i) + \frac{\lambda}{2}\|f\|_{\mathcal{H}}^2.
\end{gather}
Assuming that the loss function satisfies a simple monotonicity property, as a result of the celebrated \textit{representer theorem} \cite{Scholkopf2001-xi} we may conclude that a solution to the above problem always exists in the linear span of the set $\{k(x_i, \cdot) : i=1, \ldots, m\}$. In other words, estimating a function in an infinite dimensional space reduces to find $m$ coefficients $\alpha_1, \ldots, \alpha_m$ which parameterize the resulting solution $\hat f_m := \sum_{i=1}^m \alpha_i k(x_i, \cdot)$.

\paragraph{Online training.} Researchers have known for some time that training kernel-based learning algorithms can be prohibitively expensive when the dataset size $m$ is large; the problem is worse when the dataset is growing in size. Solving \eqref{eq:kernel_objective} naively can lead to computational cost that is at least cubic in $m$. A more scalable training procedure involves online updates to a current function estimate. A more thorough description of online kernel learning can be found in \cite{kivinen2004online} and \cite{Dekel2008-qb}, but we give a rough outline here. Let $f_t$ be the function reached at round $t$ of an iterative process,
\begin{gather}
    \textstyle f_t(x) = \rho + \sum_{i=1}^{t-1} \alpha_i k(x_i, x) .
\end{gather}
A simple gradient update with step size $\eta_t$, using the instantaneous regularized risk $R_{\textrm{reg}, \lambda}$ with respect to a single example $(x_t, y_t)$, leads to the following iterative procedure:
\begin{align}
    f_{t+1} &= f_t - \eta_t \partial_f R_{\textrm{reg}, \lambda}[f, (x_t, y_t)]|_{f=f_t} \\
    &= (1-\eta_t \lambda)f_t - \eta_t l^{\prime}(f_t(x_t), y_t)k(x_t, \cdot) \label{fifth}
\end{align}
In short: the algorithm at time $t$ maintains a set of points $x_1, \ldots, x_{t-1}$ and corresponding corresponding coefficients $\alpha_1, \ldots, \alpha_{t-1}$ and offset $\rho$, and when a new example $x_{t}$ arrives, due to (\ref{fifth}), the coefficient $\alpha_t$ is created, and the other $\alpha_i$s are scaled down:
\begin{align}
    \alpha_{t} &:= -\eta l^{\prime}(f_t(x_t), y_t), \textrm{ } \textrm{for } i=t \label{sixth}\\
    \alpha_i &\leftarrow (1-\eta\lambda)\alpha_i, \textrm{ for } i \leq t \label{seventh}
\end{align}
In our implementation, we add $n$ multiple examples at once as a minibatch at every round. For example, at round $t$, the input is not a single example $(x_t, y_t)$ but $n$ examples $(x_{n(t-1)+1}, y_{n(t-1)+1}), \cdots, (x_{nt}, y_{nt})$. Thus, we change (\ref{sixth}) and (\ref{seventh}) as below. Also, $\rho$ is updated as an average of $n$ new coefficients.
\begin{align*}
    \alpha_i &:= -\eta l^{\prime}(f_t(x_i), y_i), \textrm{ } \textrm{for }\textrm{ }n(t-1)<i \leq nt & \\
    \alpha_i &\leftarrow (1-\eta\lambda)\alpha_i, \textrm{ for } \textrm{ }i \leq n(t-1) & \rho := \textstyle \frac{1}{n}\sum_{j=n(t-1)+1}^{nt} \alpha_{j}
\end{align*}

\paragraph{Limiting the budget.} One may note that the above algorithm scales quadratically with $t$, since after a given number of $t$ rounds must compute $k(x_i, x_t)$ for all $i < t$. But this can be alleviated with a careful budgeting policy, where only a limited cache of $x_i$'s and $\alpha_i$ are stored. This is natural for a number of reasons, but especially given that $\alpha$'s decay exponentially and thus each $\alpha$ will fall below $\epsilon$ after only $\log(1/\epsilon)/(\eta\lambda)$ updates. The issue of budgeting and its relation to performance and computational issues was thoroughly explored by Dekel et al. \cite{Dekel2008-qb}, and we refer the reader to their excellent work. In our experiments we relied on the ``Remove-Oldest'' method akin to first-in-first-out (FIFO) caching. Then, at round $t$, let's say a fixed budget size of our online kernel classifier is $B$, and $w_{nt-B+1}, w_{nt-B+2}, \cdots, w_{nt}$ are key examples saved in the budget. As the result, after we finish training on the minibatch at round t, we will get a classifier function $f$ as following.
\begin{gather} \label{eq:classifier_function}
    \textstyle f_t(x)= \rho + \sum_{i=nt-B+1}^{nt} \alpha_i k(w_i, x)
\end{gather}
    
\subsection{Objective function and training}

In OKGAN, the discriminator in the original GAN formulation \eqref{eq:ganformulation} is regarded as the online kernel classifier, and it is obtained not from a parametric family of neural networks but from RKHS $\mathcal{H}$. The goal of the online kernel classifier is to separate the real data and the fake data. When we obtain the classifier function after each batch, its value of real data and fake data is respectively positive and negative. That's why we use a hinge loss while formulating a min-max objective of OKGAN. If the generator $G$ is parameterized by $\theta_g$, the objective of OKGAN is
\begin{gather} \label{okgan_minmax_obj}
    \max_{\theta_g} \min_{f \in \mathcal{H}} \mathbb{E}_{x \sim p_{\text{real}}}[\max (0, 1-f(x))] + \mathbb{E}_{z \sim p_z} 
   [\max (0, 1+f(G_{\theta_g}(z)) )] 
\end{gather}
We obtain the online kernel classifier $f$ through the process in \ref{online_kernel_classifier} after training one batch of the dataset. Then, we use the objective function of the generator $G$ as:
\begin{gather} \label{okgan_generator_loss}
    \min_{\theta_g} \Big\{ V(\theta_g)=\mathbb{E}_{z \sim p_z} [\max (0, 1-f(G_{\theta_g}(z)) )] \Big\}
\end{gather}
We use \eqref{okgan_generator_loss} as the loss function for the generator rather than $-\mathbb{E}_{z \sim p_z} 
   [\max (0, 1+f(G_{\theta_g}(z)) )]$, which is the second term in \eqref{okgan_minmax_obj} with opposite sign. It is the same reason that non-saturating loss is preferred than minimax loss \cite{fedus2017many}.

\paragraph{Objective function of OKGAN with encoder} OKGAN has superior performance on low-dimensional data such as 2d synthetic datasets (See Table \ref{table1}). But without additional representation power, it struggles to generate high-quality images that have been the hallmark of other GAN architectures. However, we find that this is remedied by adding an encoder layer. Moreover, the encoder $E$ enables us to calculate the kernel with high dimensional data such as complicated image datasets. $E$ is also a neural network and trained in a way that separates real data($x_{\textrm{real}}$) and fake data($x_{\textrm{fake}}$) because the online kernel classifier should recognize $E(x_{\textrm{real}})$ as real and  $E(x_{\textrm{fake}})$ as fake. From the perspective of OKGAN with the encoder, a combination of the encoder and the online kernel classifier is considered as the discriminator. Thus, when $G$ is parameterized by $\theta_g$ and $E$ is parameterized by $\theta_e$, we acquire an minmax objective of OKGAN as:
\begin{gather}
    \max_{\theta_g} \min_{f \in \mathcal{H}, \theta_e} \mathbb{E}_{x \sim p_{\text{real}}}[\max (0, 1-f(E_{\theta_e}(x)))] + \mathbb{E}_{z \sim p_z} 
   [\max (0, 1+f(E_{\theta_e}(G_{\theta_g}(z))) )] 
\end{gather}
There are 3 steps to train OKGAN with the encoder. First, $2N$ samples($x_1, \cdots, x_{2N}$), where N samples are real and others are fake, after passing the encoder become $E(x_1), \cdots, E(x_{2N})$, and we get an online kernel classifier $f$ from these. Second, the generator $G$ is updated based on a generator objective function with the updated $f$ and the existing $E$. Finally, the encoder $E$ is updated based on a encoder objective function with the updated $f$ and $G$. The objective function of $G$ and $E$ is shown as below. 
\begin{align}
    \min_{\theta_g} \Big\{ V(\theta_g) &= \mathbb{E}_{z \sim p_z} [\max (0, 1-f(E_{\theta_e}(G_{\theta_g}(z))))] \Big\} \\
    \min_{\theta_e} \Big\{ V(\theta_e) &= \mathbb{E}_{x \sim p_{\textrm{real}}} [\max (0, 1-f(E_{\theta_e}(x)))] + \mathbb{E}_{z \sim p_z} [\max (0, 1+f(E_{\theta_e}(G_{\theta_g}(z))))] \Big\}
\end{align}

\subsection{Flexibility on data representation through kernels}

OKGAN successfully generates classical image datasets (see Section \ref{experiment}) by achieving flexibility on data representation through the choice of kernels. We implement commonly used kernels and mixtures of them in the online kernel classifier, which are Gaussian kernel($k_\gamma^{\textrm{rbf}}$), linear kernel($k^{\textrm{linear}}$), polynomial kernel($k^{\textrm{poly}}$), rational quadratic kernel($k_\alpha^{\textrm{rq}}$), mixed Gaussian kernel($k^{\textrm{rbf}}$), and mixed RQ-linear kernel($k^{\textrm{rq}*}$) \cite{binkowski2018demystifying}. The mathematical form of kernels is:
\begin{align*}
k_\gamma^{\textrm{rbf}}(x, x^\prime) &= \exp{(-\gamma \|x-x^\prime\|^2)} & k_\alpha^{\textrm{rq}}(x, x^\prime) &= \Big( 1+\frac{\|x-x^\prime\|^2}{2\alpha}\Big)^{-\alpha} \\
\textrm{ }k^{\textrm{linear}} &= \langle x, x^\prime \rangle & k^{\textrm{rq}*}(x, x^\prime) &= k^{\textrm{linear}} + \sum_{\alpha \in \mathcal{A}} k_\alpha^{\textrm{rq}}(x, x^\prime)\\
\textrm{ }k^{\textrm{poly}}(x, x^\prime) &= (\gamma \langle x, x^\prime \rangle+r)^d &
k^{\textrm{rbf}}(x, x^\prime) &= \sum_{\gamma \in \Gamma} \exp{(-\gamma \|x-x^\prime\|^2)}\\
\end{align*}



\vspace{-3em}
\section{Experiments}
\label{experiment}

In this section, we provide experimental results of OKGANs in both quantitative and qualitative ways. We quantitatively compare OKGANs with other GANs on 2D synthetic datasets and show how well OKGANs solve the mode collapse problem, using quantitative metrics proposed earlier in \cite{srivastava2017veegan, lin2018pacgan, xiao2018bourgan}. Moreover, we analyze OKGANs qualitatively on classical image datasets and observe that OKGANs do not suffer from cycling behavior on 2D synthetic datasets, shown through our novel visualization technique.


\subsection{Experimental setup}
\label{experimental_setup}

\paragraph{Datasets} We use 2D synthetic datasets for the quantitative analysis on mode collapse, specifically, we use 2D-grid, 2D-ring, and 2D-circle~\cite{srivastava2017veegan, lin2018pacgan, xiao2018bourgan}. The 2D-grid and 2D-ring datasets are Gaussian mixtures with 25 and 8 modes, organized in a grid shape and a ring shape respectively. The specific setup of these two datasets is the same as \cite{lin2018pacgan}. The 2D-circle dataset, which is proposed in \cite{xiao2018bourgan}, consists of a continuous circle surrounding another Gaussian located in the center, and we follow the setup of \cite{xiao2018bourgan}. Additionally, MNIST, SVHN (Street View House Numbers), and CelebA are all used for the qualitative analysis. More details of datasets are in Appx. \ref{appendix_experiment}.

\paragraph{Generator \& encoder} For the quantitative analysis and cycling behavior on 2D synthetic datasets with OKGAN, we need a neural network architecture only for the generator, since the discriminator is formed by the online-kernel classifier. The generator architecture of OKGAN is the same as one of PacGAN~\cite{lin2018pacgan}, and we use the online kernel classifier instead of the discriminator of PacGAN. In experiments with classical image datasets for the qualitative analysis, we use DCGAN~\cite{radford2015unsupervised} architecture to the OKGAN. The generator of OKGAN is the same as that of DCGAN, and the encoder of OKGAN is a reverse architecture of the generator. The output dimension of the encoder is 100 for MNIST, SVHN, and CelebA. 

\paragraph{Kernel choice} The appropriate choice of a kernel is an essential part of the online kernel classifier. Since all 2D synthetic datasets are a Gaussian mixture distribution, we choose Gaussian kernel for experiments on 2D synthetic datasets. When it comes to learning a real distribution of the 2D synthetic datasets, it is significant to control the value $\gamma$ in the kernel during training OKGAN. The small $\gamma$ enables the generator to explore all different kinds of modes by smoothing the landscape of the kernel function. On the contrary, the large $\gamma$ helps fake points, which are previously located between modes, move to one of the nearest modes. Therefore, the initial $\gamma$ is small, and we increase $\gamma$ with a fixed ratio to make large in the end. The initial $\gamma$ value for 2D-grid and 2D-circle is $0.2$, and the initial $\gamma$ for 2D-ring is $3.2$. The rate of increase is same for all 2D synthetic datasets as $1.0015$.

For the qualitative analysis on classical image datasets, Gaussian kernel ($\gamma=0.01$) works well on MNIST and polynomial kernel ($\gamma=0.01, r=0, d=3$) works well on SVHN dataset. For CelebA dataset, we use the mixed gaussian kernel, where $\Gamma = \{\frac{1}{2 \times 2^2}, \frac{1}{2 \times 5^2}, \frac{1}{2 \times 10^2}, \frac{1}{2 \times 20^2}, \frac{1}{2 \times 40^2}, \frac{1}{2 \times 80^2}\}$~\cite{binkowski2018demystifying}. All coefficients of kernels in these experiments are constant during training.

\paragraph{Other hyperparameters in online kernel classifier} We use different budget size $B$ for each dataset. The budget size $B$ is 4096 for 2D synthetic datasets. In addition to this, $B$ is 700 for MNIST, 2000 for SVHN, and 1000 for CelebA. The budget size for 2D synthetic datasets is the highest because OKGANs only rely on the online kernel classifier without the encoder in this case. Moreover, there are several other hyperparameters such as regularization term($\lambda$) and step size($\eta$) and they are all constant during training. More technical details will be explained in Appx. \ref{appendix_arch_hyper}.

\paragraph{Evaluation Metrics} The evaluation metrics for 2D synthetic datasets are also previously proposed by \cite{srivastava2017veegan, lin2018pacgan}, which are \emph{\# of mode}, \emph{percentage of high-quality samples}, and \emph{reverse Kullback-Leibler(KL) divergence}. Let's say a standard deviation of Gaussian is $\sigma$, and a set of generated samples is $\{x_1, \cdots, x_N\} = \mathcal{X}$. An entire set of modes is $C=\{c_1, \cdots, c_M\}$. Then, \emph{\# of mode}, \emph{percentage of high-quality samples} is defined as:
\begin{align*}
    &\textrm{\emph{\# of mode}} = \sum_{i=1}^M\mathbbm{1} [\min_{x \in \mathcal{X}} \|c_i-x\|]<3\sigma] \\
    &\emph{percentage of high-quality samples} = 100 \times \frac{1}{N}\sum_{i=1}^N\mathbbm{1} [\min_{c \in \mathcal{C}} \|x_i-c\|]<3\sigma]
\end{align*}

 We calculate \emph{reverse KL divergence}\cite{kullback1951information} by considering real and fake distributions as discrete distributions. GAN with high \emph{\# of mode}, high \emph{percentage of high-quality samples}, and low \emph{reverse KL divergence} is regarded as good at solving mode collapse and learning real distribution well.

\subsection{Quantitative analysis}

For 2D synthetic datasets, we compare OKGAN with the two most powerful unconditional GANs on solving mode collapse, called PacGAN~\cite{lin2018pacgan} and BourGAN~\cite{xiao2018bourgan}. Since BourGAN uses different architectures, we apply neural network architectures of PacGAN to the BourGAN framework. Thus, we refer to the performance of PacGAN in \cite{lin2018pacgan} and measure the performance with new BourGAN. The quantitative performance of these three GANs on 2D-grid, 2D-ring, and 2D-circle is summarized in Table~\ref{table1}. For 2D-circle, we only compare OKGAN with BourGAN which proposes such dataset for the first time. Our results are averaged over 10 trials. More experiments with 2D synthetic datasets by changing the number of modes are shown in Appx. \ref{appendix_2d_synthetic}.

\begin{table}[!htbp]
\caption{Quantitative results on 2D synthetic datasets. For 2D-circle, at every trial, "center captured" is 1 when the center is captured and 0 when it is not captured.}
\label{table1}
\centering\footnotesize
\renewcommand\arraystretch{1.0}
\renewcommand{\tabcolsep}{3.7pt}
\begin{tabular}{crrrrrrrrr}
\toprule
\multicolumn{1}{l}{} 
& \multicolumn{3}{c}{2D-grid} 
& \multicolumn{3}{c}{2D-ring}  
& \multicolumn{3}{c}{2D-circle} \\
\multicolumn{1}{l}{} 
& \multicolumn{1}{c}{\#modes} 
& \multicolumn{1}{c}{high} 
& \multicolumn{1}{c}{reverse} 
& \multicolumn{1}{c}{\#modes} 
& \multicolumn{1}{c}{high} 
& \multicolumn{1}{c}{reverse} 
& \multicolumn{1}{c}{center} 
& \multicolumn{1}{c}{high} 
& \multicolumn{1}{c}{reverse} \\
\multicolumn{1}{l}{} 
& \multicolumn{1}{c}{(max 25)} 
& \multicolumn{1}{c}{quality(\%)} 
& \multicolumn{1}{c}{KL} 
& \multicolumn{1}{c}{(max 8)} 
& \multicolumn{1}{c}{quality(\%)} 
& \multicolumn{1}{c}{KL}
& \multicolumn{1}{c}{captured} 
& \multicolumn{1}{c}{quality(\%)} 
& \multicolumn{1}{c}{KL}  \\ \midrule
PacGAN 
& 23.8 & 91.3 & 0.13
& 7.9 & 95.6 & 0.07 
& - & - & - \\
BourGAN 
& 24.8 & \textbf{95.1} & 0.036
& 7.9 & \textbf{100.0} & 0.019
& 0.5 & \textbf{99.9} & 0.015 \\ \midrule
\textbf{OKGAN} 
& \textbf{25.0} & 86.2 & \textbf{0.006}
& \textbf{8.0} & 95.3 & \textbf{0.002} 
& \textbf{1.0} & 98.1 & \textbf{0.0003} \\ 
\bottomrule
\end{tabular}
\end{table}

\begin{figure}[!htbp]
    \centering
    \begin{subfigure}[b]{0.3\textwidth}
    \includegraphics[width=\textwidth]{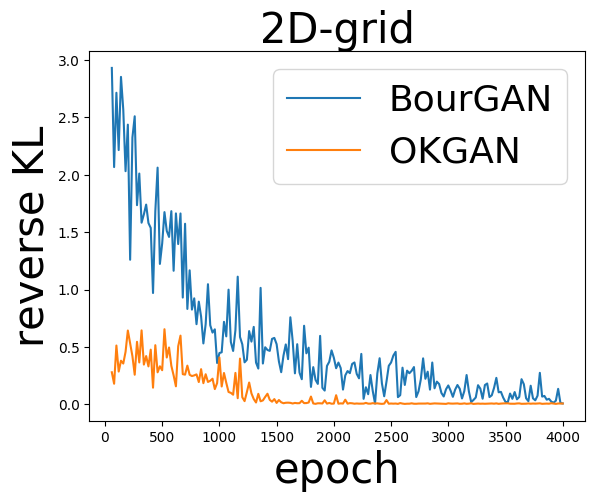}
    \end{subfigure}
    \quad
    \begin{subfigure}[b]{0.3\textwidth}
    \includegraphics[width=\textwidth]{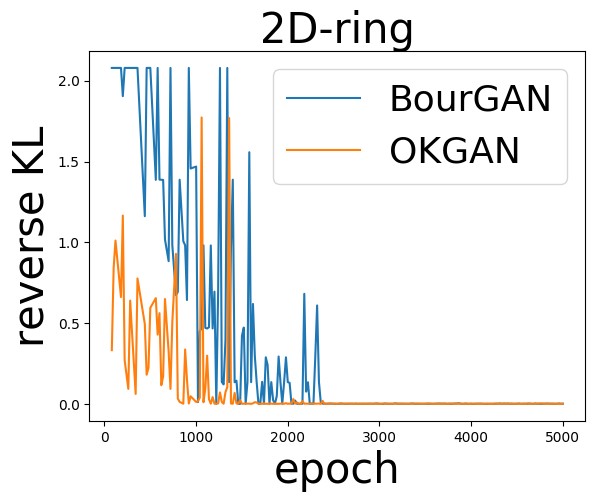}
    \end{subfigure}
    \quad
    \begin{subfigure}[b]{0.3\textwidth}
    \includegraphics[width=\textwidth]{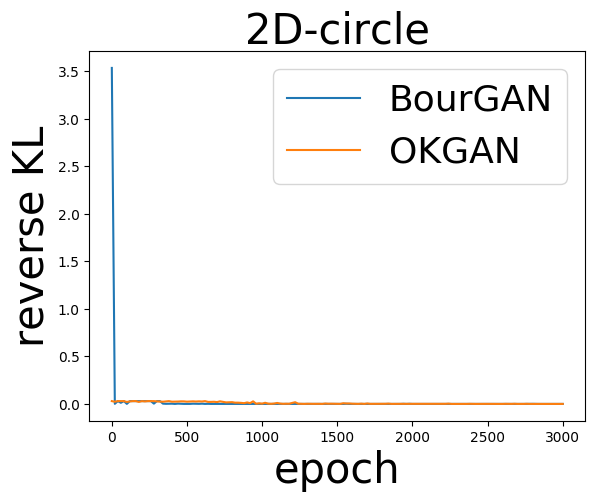}
    \end{subfigure}
    \caption{Reverse KL divergence graph on 2D-grid(Left), 2D-ring(Middle), and 2D-circle(Right).}
    \label{synthetic_reverse_kl_graph}
\end{figure}

As you can see from Table \ref{table1} and Appx. \ref{appendix_2d_synthetic}, OKGAN shows the best performance overall in terms of mitigating mode collapse. In contrast with other GANs, OKGAN captures all modes for all 2D synthetic datasets. The remarkable point is that \emph{reverse KL divergence} of OKGAN is the lowest for all three datasets. This indicates that OKGAN not only produces all modes but also generates each mode with the similar proportion in a real distribution. Furthermore, the fake distribution of OKGAN converges to the real distribution faster than that of BourGAN, and OKGAN is trained in a more stable way (See Figure \ref{synthetic_reverse_kl_graph}). Therefore, we conclude that OKGAN increases the diversity of generated samples by taking advantage of a kernel-based non-parametric discriminator.


%

\subsection{Qualitative analysis}
\label{sec:qualitative}
\begin{figure}[!htbp] 
    \centering
    \begin{subfigure}[b]{0.35\textwidth}
    \includegraphics[width=\textwidth]{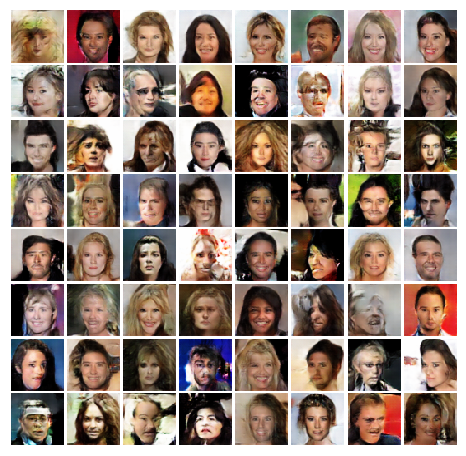}
    \caption{DCGAN}
    \end{subfigure}
    \quad
    \begin{subfigure}[b]{0.35\textwidth}
    \includegraphics[width=\textwidth]{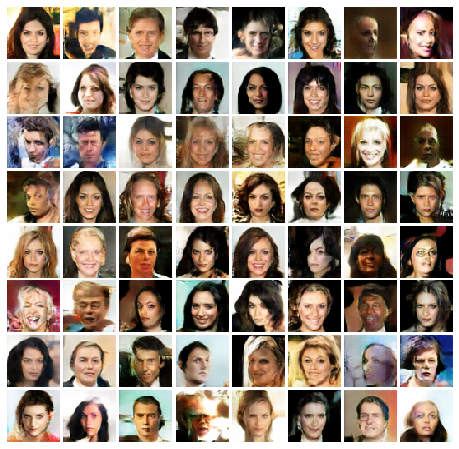}
    \caption{OKGAN}
    \end{subfigure}
    \caption{Qualitative comparison on CelebA dataset}
    \label{celeba_image}
\end{figure}

\paragraph{Classical image datasets} We qualitatively compare OKGAN with DCGAN on the CelebA dataset (see Figure \ref{celeba_image}). Both DCGAN and OKGAN successfully generate fake images of celebrities. Further qualitative comparison on MNIST and SVHN are provided in Appx. \ref{appendix_qualitative}. 
\paragraph{Cycling behavior} In Figure \ref{cycling_graph}(a), we can clearly observe that Vanilla GAN (VGAN) shows the cycling behavior during training, which means the discriminator does cycle and fails to give meaningful information to the generator. In a parameter-based alternative update framework such as VGAN, it is challenging for the discriminator to chase the transitions of the generator with a slow pace of parameter updates. However, in the case of OKGAN, by obtaining a closed-form discriminator with a non-parametric kernel method, the discriminator is updated in a more aggressive fashion and separates real and fake data more effectively at every update. As you can see in Figure \ref{cycling_graph}(b), the discriminator of OKGAN tends to find the optimal discriminator with no apparent cycling behavior, which also leads to solving the mode collapse problem in the end.

\section{Discussion}


In this work, we propose OKGAN, a new type of GAN whose discriminator contains the online kernel classifier. We provide a novel method for visualizing the cycling behavior of GANs and empirically show that OKGAN does not suffer from this issue. Moreover, with a kernel-based non-parametric discriminator, OKGAN successfully learns 2D synthetic data with no mode collapse and generates high quality samples in image datasets. In future, the deeper theoretical understanding on dynamics of GANs with the non-parametric discriminator can be discussed. Applying the idea of combining the kernel method with neural networks \cite{wilson2016deep} to GANs will be another interesting future work.  





\bibliography{neurips_2020} 
\bibliographystyle{IEEEtranS}

\clearpage
\appendix

\section{Algorithm \& training details}
\label{appendix_alg_training}

\begin{algorithm}
\caption{OKGAN with encoder}\label{okgan_alg}
\begin{algorithmic}
\Require{$\alpha$ learning rate, $\gamma$ learning rate decay, $N_b$ batch size, $n_b$ batch size of mini-batch, $B$ budget size, $n_g$ number of iterations of generator per discriminator update.}
\While{not converged}
\State Sample $\{x_i\}_{i=1}^{N_b} \sim p_{\textrm{real}}$ \& $\{z_i\}_{i=1}^{N_b} \sim p_z$
\If{not first iteration}
\State $g_{\theta_e} \leftarrow \nabla_{\theta_e} \frac{1}{N_b} \sum_{i=1}^{N_b} \max (0, 1-f(E_{\theta_e}(x_i))) + \frac{1}{N_b} \sum_{i=1}^{N_b} \max (0, 1+f(E_{\theta_e}(G_{\theta_g}(z_i))))$
\State $\theta_e \leftarrow \theta_e - \alpha \cdot \textrm{Adam} (\theta_e, g_{\theta_e})$ 
\EndIf
\State $2N_b$ inputs for the online kernel classifier are $E(x_1), \cdots, E(x_{N_b}), E(G(z_1)), \cdots, E(G(z_{N_b}))$
\State In the perspective of the classifier, consider $2N_b$ as a total size of data and $n_b$ as a batch size.
\State $f \leftarrow \rho + \sum_{i=1}^B \alpha_i k(w_i, x)$
\Comment{refer to the process in Section \ref{online_kernel_classifier}}
\State $\alpha_i$s and $w_i$s are respectively coefficients and key examples saved in the current budget 
\For{$t = 1, \cdots, n_g$}
\State Sample $\{z_j\}_{j=1}^{N_b} \sim p_z$
\State $g_{\theta_g} \leftarrow \nabla_{\theta_g} \frac{1}{N_b} \sum_{j=1}^{N_b} \max (0, 1-f(E_{\theta_e}(G_{\theta_g}(z_j))))$
\State $\theta_g \leftarrow \theta_g - \alpha \cdot \textrm{Adam} (\theta_g, g_{\theta_g})$ 
\EndFor
\State $\alpha \leftarrow \alpha\gamma$
\EndWhile

\end{algorithmic}
\end{algorithm}

The above algorithm applies to OKGAN with encoder. We use OKGAN with encoder on datasets such as MNIST, SVHN, and CelebA. For the experiment on 2D synthetic datasets, we use OKGAN which contains only the generator and the online kernel classifier. The learning rate $\alpha$ and the learning rate decay $\gamma$ is respectively $5\cdot 10^{-4}$ and $0.999$ for OKGAN on 2D synthetic datasets. When it comes to dealing with classical image datasets, the learning rate and the learning rate decay is respectively $0.0002$ and $1$. Adam optimizer with $\beta_1=0.9, \beta_2=0.999$ is used for 2D synthetic datasets, and Adam optimizer with $\beta_1=0.5, \beta_2=0.999$ is used for MNIST, SVHN, and CelebA. The batch size of mini-batch $n_b$ for the online kernel classifier is $64$ for all datasets. Each batch size $N_b$ of 2D synthetic datasets, MNIST, SVHN, and CelebA is 500, 200, 128, 128.

We need to update the generator several times per one discriminator update because the generator requires many updates in order to fool the discriminator, which has a strong discriminative power by taking advantage of non-parametric kernel learning. Each number of iterations of the generator per one discriminator update $n_g$ for 2D synthetic datasets, MNIST, SVHN, and CelebA is 5, 10, 1, 3. 

\section{Architectures \& hyperparameters}
\label{appendix_arch_hyper}

\subsection{Further details of neural network architectures}

As we mention in Section \ref{experimental_setup}, we use neural network architectures of PacGAN\cite{lin2018pacgan} to BourGAN and our proposed OKGAN in the experiment with 2D synthetic datasets. Specifically, the generator has four hidden layers, batch normalization, and 400 units with ReLU activation per hidden layer. The input noise for the generator is a two dimensional Gaussian whose mean is zero, and covariance is identity. Also, the discriminator of PacGAN and BourGAN has three hidden layers with LinearMaxout with 5 maxout pieces and 200 units per hidden layer. Batch normalization is not used in the discriminator. Additionally, when we use PacGAN, we set the number of packing to be two. OKGAN does not need the encoder architecture for the experiment with 2d synthetic datasets. 

In terms of training OKGAN with classical image datasets such as MNIST, SVHN, and CelebA, we need the encoder architecture, which is a reverse of the generator. We apply DCGAN \cite{radford2015unsupervised} neural network architectures to OKGAN. The generator of OKGAN is a series of strided two dimensional convolutional transpose layers, each paired with a 2d batch norm and ReLU activation. (\# of input channel, \# of output channel, kernel size, stride, padding) are important factors for convolutional layers and convolutional transpose layers. For MNIST and CelebA dataset, we use 5 convolutional transpose layers sequentially as (100, 512, 4, 1, 0), (512, 256, 4, 2, 1), (256, 128, 4, 2, 1), (128, 64, 4, 2, 1), (64, $n_c$, 4, 2, 1). $n_c$ is 1 for MNIST and 3 for CelebA. Additionally, for MNIST, we find out that (100, 1024, 4, 1, 0), (1024, 512, 4, 2, 1), (512, 256, 4, 2, 1), (256, 128, 4, 2, 1), (128, 1, 4, 2, 1) also works well. For SVHN dataset, we firstly use fully connected layer before applying convolutional transpose layers. Then, we sequentially use 3 convolutional transpose layers as (128, 64, 4, 2, 1), (64, 32, 4, 2, 1), (32, 3, 4, 2, 1). Furthermore, the encoder of OKGAN is a series of strided two dimensional convolutional layers, each paired with a 2d batch norm and LeakyReLU activation. It is easy to figure out (\# of input channel, \# of output channel, kernel size, stride, padding) of the encoder because the encoder is simply a reverse architecture of the generator.

\subsection{Hyperparameters in the online kernel classifier}

The parameters such as types of kernels, budget size, regularization term, and step size are already discussed in Section \ref{experimental_setup}. The regularization term($\lambda$) is 0.1, and the step size($\eta$) is 0.05 for all experiments. Moreover, the online kernel classifier allows two types of the loss function; hinge loss and logistic loss. We fix the margin value as 1.0 when we use hinge loss for the online kernel classifier. Also, we fix the degree as 3 and the coef0 as 0.0 when we use the polynomial kernel. 

\section{Experiment details}
\label{appendix_experiment}

\subsection{Experiment details on 2D synthetic datasets \& further experiments}
\label{appendix_2d_synthetic}

For 2D-grid and 2D-ring, we follow the experiment setup used in \cite{lin2018pacgan}. The standard deviation of each Gaussian in 2D-grid is 0.05, and a grid value of four edges is (-4, -4), (-4, 4), (4, -4), (4, 4). In the case of 2D ring, the standard deviation of each Gaussian is 0.01, and the radius of a ring shape is 1. In addition to this, for 2D-circle, we follow the experiment setup in \cite{xiao2018bourgan}. 100 Gaussian distributions are on a circle with a radius 2, and three identical Gaussians are located at the center of the circle. The standard deviation of each Gaussian is 0.05. We generate 2500 samples from the trained generator to quantitatively compare GANs on evaluation metrics in Section \ref{experimental_setup}. 
Also, We train the model with 4000 epochs, 5000 epochs, 3000 epochs respectively for 2D-grid, 2D-ring, 2D-circle.

\begin{figure}[!htbp]
    \centering
    \includegraphics[width=0.6\textwidth]{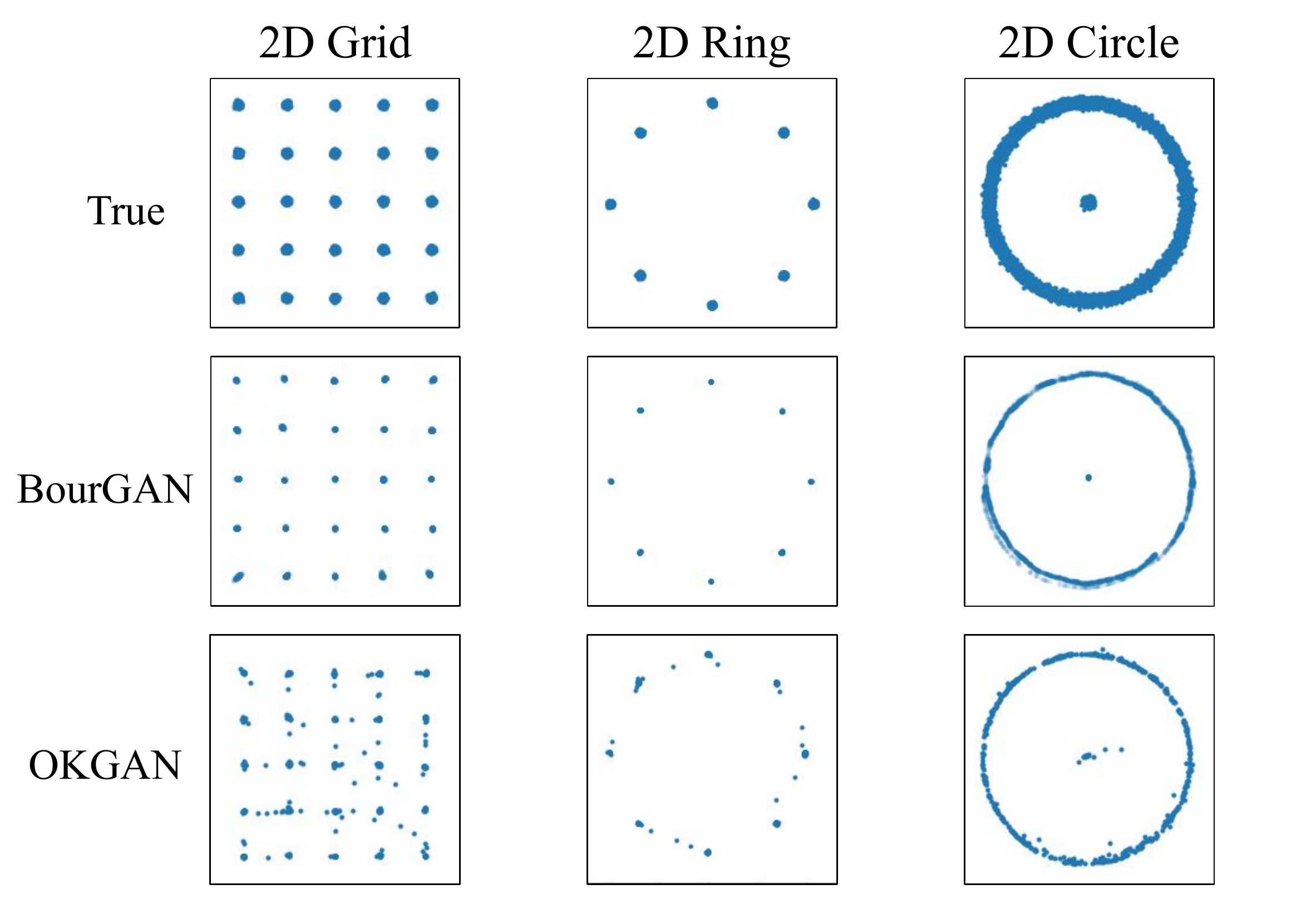}
    \caption{Qualitative comparison on 2D synthetic datasets. (OKGAN vs. BourGAN). The advantage of BourGAN is that BourGAN generates high-quality samples by avoiding unwanted samples between modes \cite{xiao2018bourgan}. In contrast, OKGAN is better at capturing modes with similar proportion in real distribution.}
    \label{2d_synthetic_comparison}
\end{figure}

In Figure \ref{2d_synthetic_comparison}, we provide the qualitative analysis of BourGAN and OKGAN on 2D-grid, 2D-ring, and 2D-circle. Moreover, we perform an additional experiment on the 2D-grid dataset with 49 modes, which shows the superiority of OKGAN in terms of achieving diversity. For this new dataset, the standard deviation of each Gaussian is 0.05, and a grid value of four edges is (-4, -4), (-4, 4), (4, -4), (4, 4). In this case, we use the initial $\gamma$ value of Gaussian kernel as 0.5. In Table \ref{table2}, only OKGAN successfully generates all 49 modes with the lowest reverse KL divergence. Every evaluation metric value is averaged over 5 trials.

\begin{table}[!htbp]
\caption{Quantitative results on 2D-grid dataset with 49 modes.}
\label{table2}
\centering\footnotesize
\renewcommand\arraystretch{1.0}
\renewcommand{\tabcolsep}{3.7pt}
\begin{tabular}{crrr}
\toprule
\multicolumn{1}{l}{} 
& \multicolumn{3}{c}{2D-grid} \\

\multicolumn{1}{l}{} 
& \multicolumn{1}{c}{\#modes} 
& \multicolumn{1}{c}{high} 
& \multicolumn{1}{c}{reverse} \\

\multicolumn{1}{l}{} 
& \multicolumn{1}{c}{(max 49)} 
& \multicolumn{1}{c}{quality(\%)} 
& \multicolumn{1}{c}{KL} \\ \midrule


PacGAN 
& 45.0 & \textbf{83.8} & 0.364 \\

BourGAN 
& 42.0 & 68.2 & 0.253 \\ \midrule

\textbf{OKGAN} 
& \textbf{49.0} & 74.3 & \textbf{0.033} \\

\bottomrule
\end{tabular}
\end{table}

\begin{figure}[!htbp]
    \centering
    \includegraphics[width=0.7\textwidth]{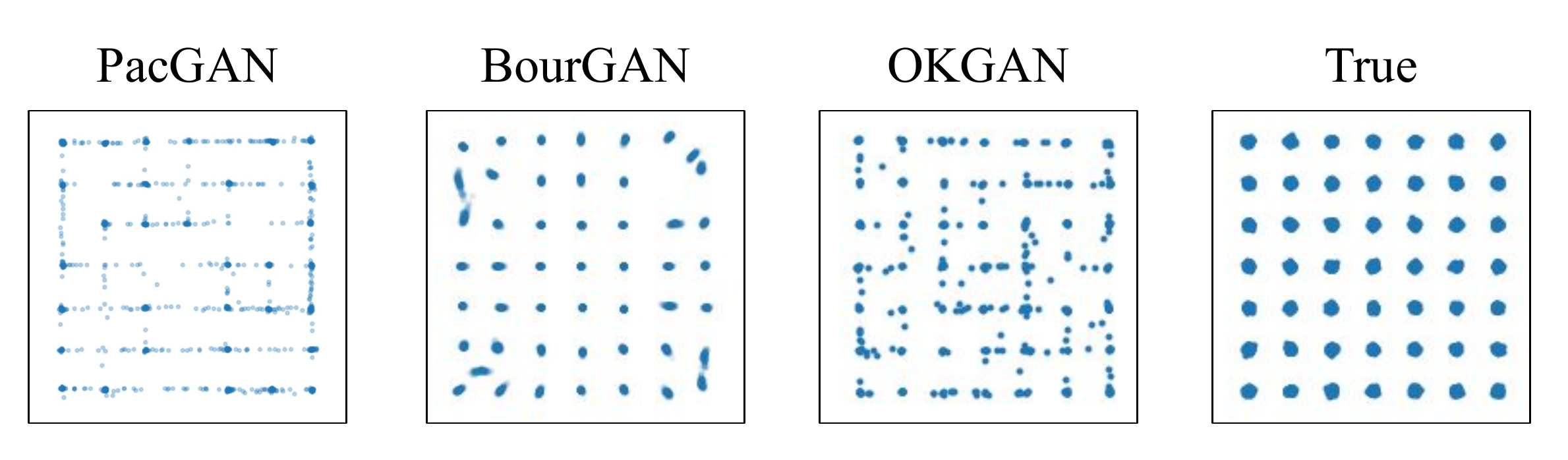}
    \caption{Qualitative comparison on 2D-grid dataset with 49 modes.}
    \label{2d_grid_49modes}
\end{figure}

\subsection{More qualitative results}
\label{appendix_qualitative}

In addition to the experiment on CelebA dataset in Section \ref{sec:qualitative}, we provide more qualitative results on other images datasets such as MNIST and SVHN. For MNIST, we use 5 convolutional transpose layers as (100, 512, 4, 1, 0), (512, 256, 4, 2, 1), (256, 128, 4, 2, 1), (128, 64, 4, 2, 1), (64, 1, 4, 2, 1) both for the generator of DCGAN and OKGAN. Then, we check how OKGAN works well on even more complicated dataset like SVHN, which contains random street numbers with some colors. We observe that OKGANs successfully generate high-quality samples both on MNIST and SVHN. 

\begin{figure}[!htbp]
    \centering
    \begin{subfigure}[b]{0.3\textwidth}
    \includegraphics[width=\textwidth]{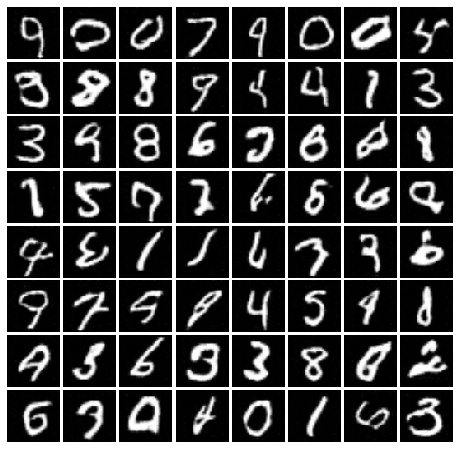}
    \caption{DCGAN}
    \end{subfigure}
    \quad
    \begin{subfigure}[b]{0.3\textwidth}
    \includegraphics[width=\textwidth]{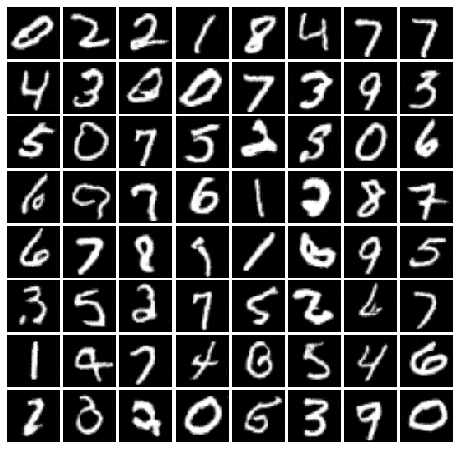}
    \caption{OKGAN}
    \end{subfigure}
    \caption{Qualitative comparison on MNIST dataset}
    \label{mnist_image}
\end{figure}

\begin{figure}[!htbp]
    \centering
    \begin{subfigure}[b]{0.3\textwidth}
    \includegraphics[width=\textwidth]{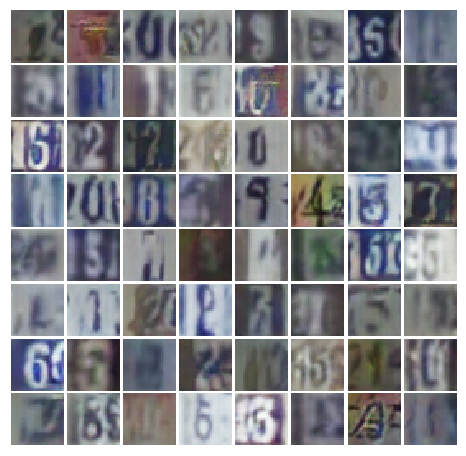}
    \caption{DCGAN}
    \end{subfigure}
    \quad
    \begin{subfigure}[b]{0.3\textwidth}
    \includegraphics[width=\textwidth]{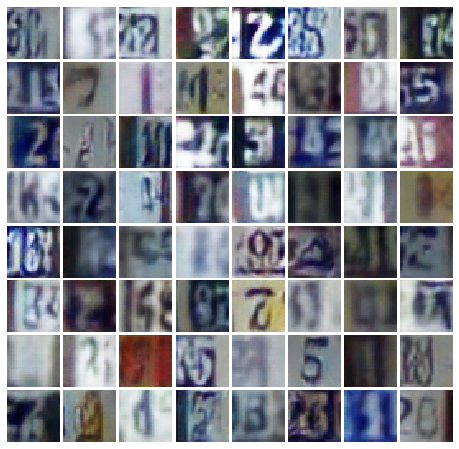}
    \caption{OKGAN}
    \end{subfigure}
    \caption{Qualitative comparison on SVHN dataset}
    \label{svhn_image}
\end{figure}

\subsection{Computational complexity analysis}

\begin{figure}[!htbp]
    \centering
    \includegraphics[width=0.5\textwidth]{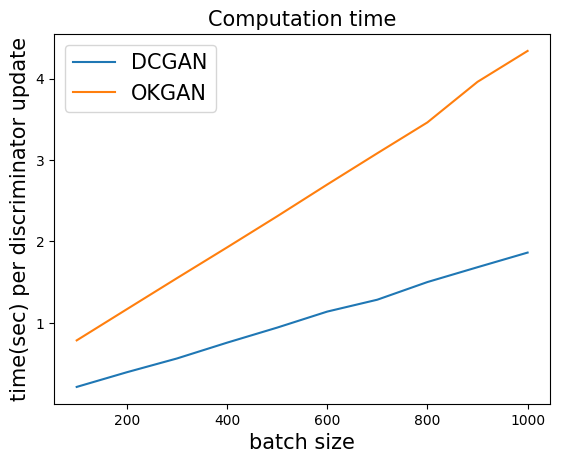}
    \caption{Computation time graph of DCGAN and OKGAN}
    \label{time_complexity_graph}
\end{figure}

We conduct computational complexity analysis on DCGAN and OKGAN with respect to the batch size. We use CelebA dataset on this experiment, and the time spent per discriminator update is measured while we increase the batch size. In this case, the discriminator of OKGAN is regarded as the combination of the encoder and the online kernel classifier. When the batch size is $S_B$, the time complexity of each update of typical GANs is $O(S_B)$ \cite{li2017mmd}. The time complexity of training the encoder of OKGAN is $O(S_B)$ as well. For the online kernel classifier, since the budget size $B$ and the batch size of mini-batch $n_b$ are fixed, the computation time of training the classifier on one mini-batch increases linearly with the batch size $S_B$. Therefore, the time complexity of the discriminator update of OKGAN is $O(S_B)$ (See Figure \ref{time_complexity_graph}). Even though we add the online kernel classifier to achieve the diversity of generated samples, OKGAN is not too computationally expensive compared to typical GANs.

\end{document}